\documentclass[letterpaper, 10 pt, conference]{ieeeconf}  

\IEEEoverridecommandlockouts                              

\usepackage{multicol}
\usepackage[bookmarks=true]{hyperref}
\usepackage{xcolor}
\usepackage{amsmath}
\usepackage{graphicx}
\usepackage{amssymb}
\usepackage[detect-weight=true, detect-family=true]{siunitx}
\usepackage[capitalize]{cleveref}
\usepackage{flushend}

\usepackage[natbib=true,
    style=ieee,
    citestyle=numeric-comp,
    backend=biber,
    maxbibnames=5,
    maxcitenames=99,
    doi=false,
    url=false,
    giveninits=true]{biblatex}
\DeclareSourcemap{
  \maps[datatype=bibtex]{
    \map{
      \step[fieldsource=url,
        match=\regexp{\\_},
        replace=\regexp{_}]
    }
  }
}

\AtBeginBibliography{\footnotesize}
\newbibmacro{string+url}[1]{%
  \iffieldundef{url}{%
  \iffieldundef{doi}{#1}{\href{http://dx.doi.org/\thefield{doi}}{#1}}}{\href{\thefield{url}}{#1}}}
\DeclareFieldFormat*{title}{\usebibmacro{string+url}{``#1''}}

\newcommand{\CF}{\text{CF}}
\newcommand{\MC}{\text{MC}}

\DeclareMathOperator*{\argmin}{arg\,min} 

\title{\LARGE \bf
Lighthouse Positioning System: Dataset, Accuracy, and Precision for UAV Research
}
\author{Arnaud Taffanel, Barbara Rousselot, Jonas Danielsson, Kimberly McGuire, Kristoffer Richardsson,\\ Marcus Eliasson, Tobias Antonsson, and Wolfgang Hönig
\thanks{All authors are with Bitcraze AB, Sweden.}%
\thanks{Email: {\tt\footnotesize firstname@bitcraze.io} or {\tt\footnotesize all@bitcraze.io}}%
\thanks{Dataset and code are available at: {\footnotesize\url{https://github.com/bitcraze/positioning_dataset}}.}%
}%

\begin{document}
\maketitle
\thispagestyle{empty}
\pagestyle{empty}

\begin{abstract}
The Lighthouse system was originally developed as tracking system for virtual reality applications. Due to its affordable price, it has also found attractive use-cases in robotics in the past.
However, existing works frequently rely on the centralized official tracking software, which make the solution less attractive for UAV swarms.
In this work, we consider an open-source tracking software that can run onboard small Unmanned Aerial Vehicles (UAVs) in real-time and enable distributed swarming algorithms.
We provide a dataset specifically for the use cases i) flight; and ii) as ground truth for other commonly-used distributed swarming localization systems such as ultra-wideband.
We then use this dataset to analyze both accuracy and precision of the Lighthouse system in different use-cases.
To our knowledge, we are the first to compare two different Lighthouse hardware versions with a motion capture system and the first to analyze the accuracy using tracking software that runs onboard a microcontroller.
\end{abstract}

\section{Introduction}


Research and development for aerial swarms often require reproducible laboratory settings, at least in the early stages.
Frequently, a motion capture system is used to mimic GPS indoors~\cite{crazyswarm}.
However, these positioning systems are quite expensive and introduce a centralized bottleneck, which make them difficult to use in classrooms, low-budget labs, or for research on fully distributed and resilient systems.
The Valve Lighthouse system (LH) was first introduced to revolutionize virtual reality, but has also found its use in robotics~\cite{DronOS,DBLP:conf/fusion/GreiffRB19,sletten2017automated}. It is about one order of magnitude cheaper than motion capture systems, easier to transport, and allows fully distributed operation as robots can compute their positions without a central computer.

When flying many robots in a small volume, the accuracy of the position tracking becomes very important.
The accuracy of the LH has been determined before using the official bulky tracking device and centralized software~\cite{niehorster2017accuracy,DBLP:journals/access/IkbalRZ21,DBLP:conf/plans/WuPBCGJZ20,9234740,luckett2018quantitative,DBLP:journals/sensors/VeenBPFT19}.
We are interested in quantifying the performance using small custom trackers (similar to~\cite{DBLP:conf/huc/HonnetL19}) with open-source software (similar to~\cite{DBLP:conf/iros/BorgesSCSV18}).
In contrast to existing work, our open-source software is designed to run in real-time onboard small UAVs --- we rely on the Crazyflie 2.1.\footnote{\url{https://store.bitcraze.io/products/crazyflie-2-1}} 
To this end, we provide a dataset that includes Lighthouse data (raw and onboard processed), motion capture ground truth in different scenarios (stationary, manual movement, flight) using the two available hardware versions of LH.
We also contribute our analysis of the data to quantify the precision (also known as jitter) and accuracy of the system.

\section{Lighthouse positioning system}\label{sec:LH}                     
In this section we first introduce the hardware of the Lighthouse (SteamVR) base stations\footnote{\url{https://www.vive.com/eu/accessory/base-station2/}, ... /basestation/} and the Lighthouse sensor board\footnote{\url{https://www.bitcraze.io/products/lighthouse-positioning-deck/}}. Afterwards, we specify the Lighthouse positioning system algorithms for estimating the Crazyflie's position by using the crossing beam method (C.B.) or by using an extended Kalman filter (EKF).

\begin{figure}[hbt!]
    \centering
    \includegraphics[width=\linewidth]{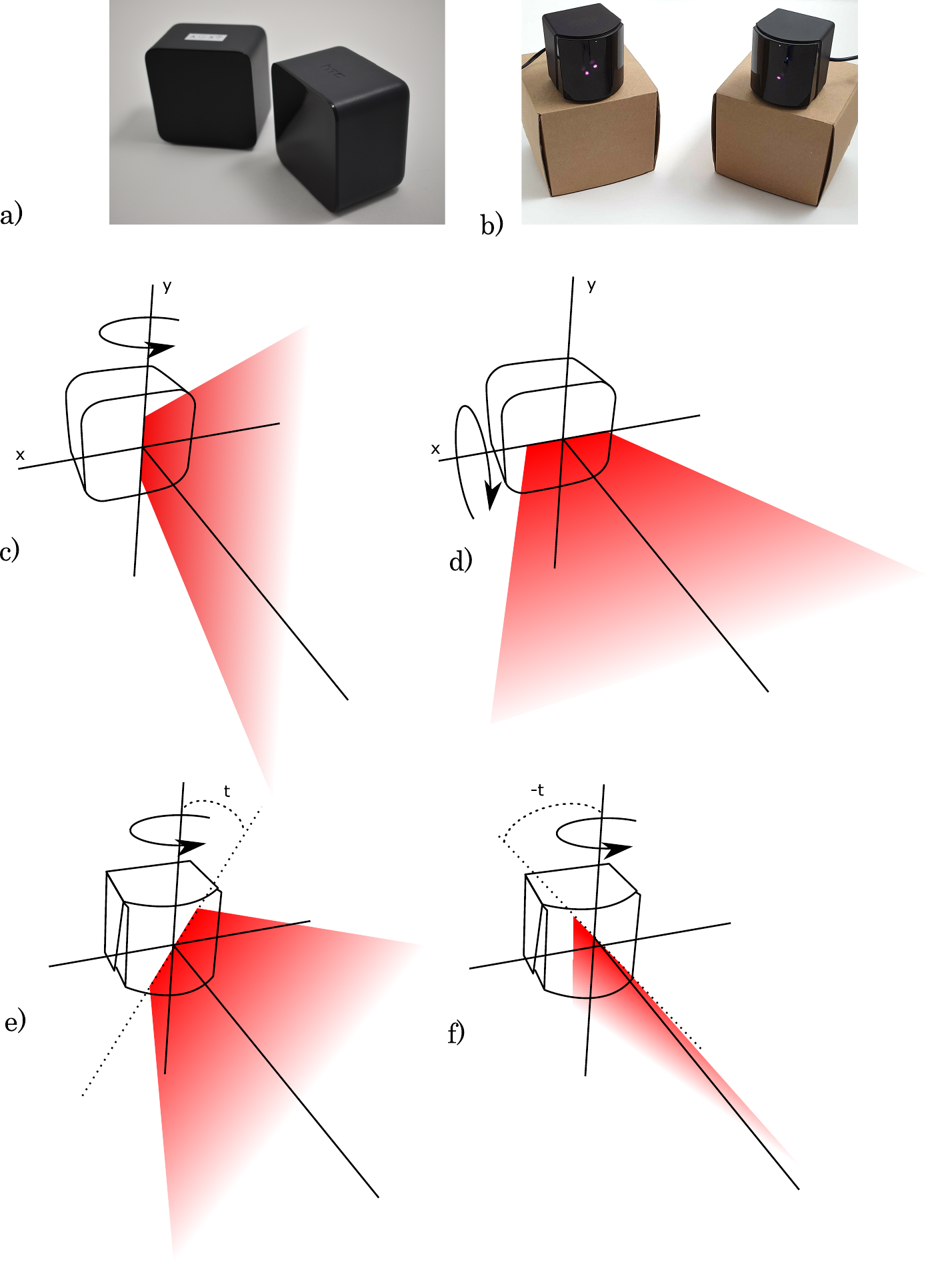}
    \caption{Pictures and operational principle of the two Lighthouse positioning system versions. LH1 is shown in a) and has two rotating drums creating IR light planes that are perpendicular to each other c) and d). LH2 is shown in b) and has just one rotating drum with the two sweep planes angled at a tilt angle $t$ shown in e) and f).}
    \label{fig:sweep_angles}
\end{figure}

\subsection{Hardware}

The Lighthouse base stations contain drums which rotate infrared (IR) light planes that are captured by the IR receivers on the Lighthouse sensor board mounted on the robot (so-called deck). Currently, there are two versions:

\begin{itemize}
    \item V1 (\cref{fig:sweep_angles}a) has two rotating drums with each carrying one sweep (\cref{fig:sweep_angles}c and d). The rotating drums of two base stations are synced with each other by a sync cable or visual line of sight. This paper will refer to this version as LH1.
    \item V2 (\cref{fig:sweep_angles}b) has one rotating drum with two planes each at a different angle (\cref{fig:sweep_angles}e and f). Depending on the channel configuration of the base station, the drums rotate at a slightly different rate. Thus, light planes of different base stations may periodically collide. This paper will refer to this as LH2.
\end{itemize}

\Cref{tab:basestation_characteristics} shows the the tilt angles ($t_{p}$) of the IR light planes and the drum's rotation matrices ($R_{d}$) for both LH1 and LH2.

\begin{table}[hbt!]
\caption{Characteristics of LH1 and LH2 with $I$ being an identity matrix.}
\label{tab:basestation_characteristics}
\centering
\begin{tabular}{ c || c | c || c | c}
 Version: & \multicolumn{2}{c||}{LH1} &  \multicolumn{2}{c}{LH2} \\ 
Light Plane: & $1^{\text{st}}$ &  $2^{\text{nd}}$ &  $1^{\text{st}}$ & $2^{\text{nd}}$\\
  \hline
  \hline
 \(t_{p}\) & 0 & 0 & $-\pi/6$ & $\pi/6$  \\
$R_{d}$	& $I$ & \tiny{$\begin{bmatrix}
1 & 0 & 0\\
0 & 0 & 1\\
0& -1& 0
\end{bmatrix}$} &  $I$  &  $I$ 
\end{tabular}
\end{table}

\begin{figure}[hbt!]
    \centering
    \includegraphics[width=0.5\linewidth]{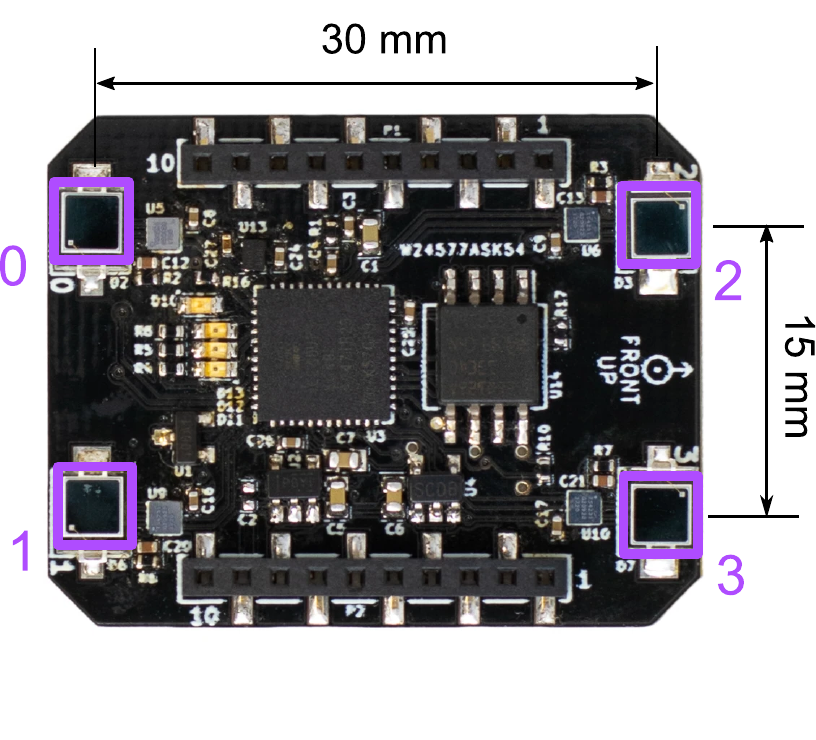}
    \caption{Lighthouse deck with 4 TS4231 IR receivers.}
    \label{fig:lhdeck}
\end{figure}

The Lighthouse deck, see \cref{fig:lhdeck}, contains 4 sensors that are able to detect passing IR light planes from both LH1 and LH2. The received signals are decoded by an FPGA,  which transmits the timestamps, pulse-width, and, for LH2, the decoded base station synchronization time to the Crazyflie. On the STM32F4 chip of the Crazyflie, the data is processed to plane sweep angles, which are used for the position estimation either by the C.B. method (\cref{sec:crossingbeam}) or the EKF (\cref{sec:measurement_model}). There are also calibration models for LH1 and LH2 that are implemented in the firmware of the Crazyflie\footnote{\url{https://github.com/bitcraze/crazyflie-firmware}}, which will not be covered in detail in this paper.

\begin{figure}[hbt!]
    \centering
    \includegraphics[width=\linewidth]{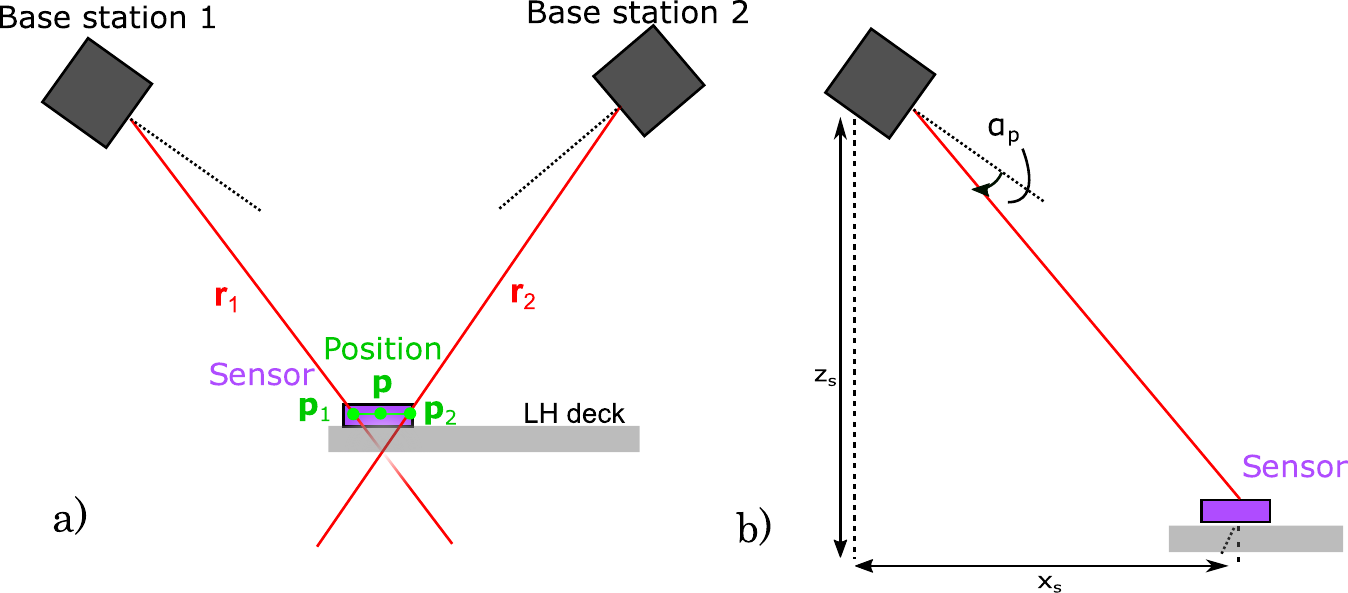}
    \caption{The two position estimation methods, where a) depicts the crossing beam method (C.B.) and b) depicts state variables necessary for the EKF's sweeping beam measurement model.  }
    \label{fig:crossing_beam}
\end{figure}

\subsection{Crossing Beam Method}
\label{sec:crossingbeam}

The crossing beam method is developed by the open-source community\footnote{\url{https://github.com/ashtuchkin/vive-diy-position-sensor/}} for LH1 and has been extended by us to work for LH2 on the Crazyflie. The method needs two base stations to be in direct line of sight with the Lighthouse deck on the Crazyflie (see \cref{fig:crossing_beam}a). The intersection of the two light planes of the base station (\(b\)) to one of the Lighthouse deck's sensor ($s$) results in a ray (\(\pmb r_{b,s}'\)) that points towards the Lighthouse deck. The ray's origin and direction are then converted to the global coordinate system (\(\pmb r_{b,s}\)) by using the rotation matrix ($R_b$) and translation ($\pmb t_b$) of the base station, i.e., with slight abuse of notation:
\begin{equation}
    \pmb r_{b, s} = R_b \cdot  \pmb r_{b, s}' + \pmb t_b.
\end{equation}

Thus, the rays of the two base stations that are perceived by sensor $s$ in global coordinates are $\pmb r_{1, s}$ and $\pmb r_{2, s}$.
The position $\pmb p_s = (x_s, y_s, z_s)^\top$ of sensor $s$ is then calculated by first solving\footnote{\url{http://geomalgorithms.com/a07-_distance.html\#Distance-between-Lines}} the following optimization problem
\begin{equation}
     \pmb p_{1, s}, \pmb p_{2, s} = \argmin_{\pmb  p_{1}\in\pmb  r_{1, s},~\pmb  p_{2}\in \pmb r_{1, s} } \|\pmb p_{1} - \pmb p_{2}\|_2
\end{equation}
to compute the two points $\pmb p_{1,s}$ and $\pmb p_{2,s}$ on the two rays that are the closest to each other. Then, we compute $\pmb p_s$ and estimate the error $\delta$ using
\begin{equation}
    \pmb p_s = \frac{\pmb  p_{1, s}+\pmb  p_{2, s}}{2} \text{ and } \delta = \|\pmb p_{1, s} - \pmb p_{2, s}\|^2,
\end{equation}
respectively. Here, the position of the sensor ($\pmb p_s$) is estimated by using the middle between the two closest points. Afterwards, the average of the estimated position of all the sensors ($s=0 \dots 3$) is used as a measured position. For our analysis, we rely on the estimated error ($\delta$) to filter the measured data.

\subsection{EKF Lighthouse Measurement model}\label{sec:measurement_model}

When using the EKF implementation on the Crazyflie~\cite{MuellerCovariance2016}, each IR light plane {$p$} can be used as a measurement. This enables the Crazyflie to fly with only one base station. This section explains the measurement model that is used for our implementation.


The observation model (see \cref{fig:crossing_beam}b) maps the relevant part of the EKF state, the estimated $x_{s}$, $y_{s}$, and $z_{s}$ components of the sensor position in the global coordinate frame, to the measured observations $\alpha_{p}$:
\begin{align}
    \alpha_{p} = \arctan\dfrac{y_{s}}{x_{s}} + \arcsin\dfrac{z_{s}\tan t_{p}}{r_s},\\
  \text{where } r_{s} = \sqrt{x_s^2 + y_s^2}.
\end{align}

The measurement model is linearized by the Jacobian:
\begin{align}
    \pmb g_{p} = \left(\dfrac{\partial\alpha_{p}}{\partial x_s}, \dfrac{\partial\alpha_{p}}{\partial y_s}, \dfrac{\partial\alpha_{p}}{\partial z_s}\right)\\
    \pmb g_{p} = \left(\frac{-y_s-x_sz_sq_{p}}{r_s^2},~\frac{x_s-y_sz_sq_{p}}{r_s^2},~q_{p}\right)\\
     \text{with: } q_{p}~=~\frac{\tan t_{p}}{\sqrt{r_s^2 - ( z_s\tan t_{p})^2}}
\end{align}

This Jacobian is first rotated to the coordinate system of the base stations with the drum's rotation matrix ($R_{d}$, see \cref{tab:basestation_characteristics}) and the rotation matrix of the orientation of the base station itself ($R_{b}$):
\begin{equation}
     \pmb g_{p}' = (R_{d}\cdot R_{b}^{-1})^{-1}\cdot \pmb g_{p} =  R_{b}\cdot R_{d}^{-1}\cdot \pmb g_{p}.
\end{equation}

With the measurement model, the characteristics of \cref{tab:basestation_characteristics}, and the measured sweep angles of each sensor,  the EKF is able to estimate the position of the Crazyflie.

\section{Data Collection Setup}

In this section we introduce our flight area configuration with the LH base stations and the ground truth measurement, as well as the hardware configuration of the Crazyflie used for the data collection.

\begin{figure}[hbt!]
    \centering
    \includegraphics[width=\linewidth]{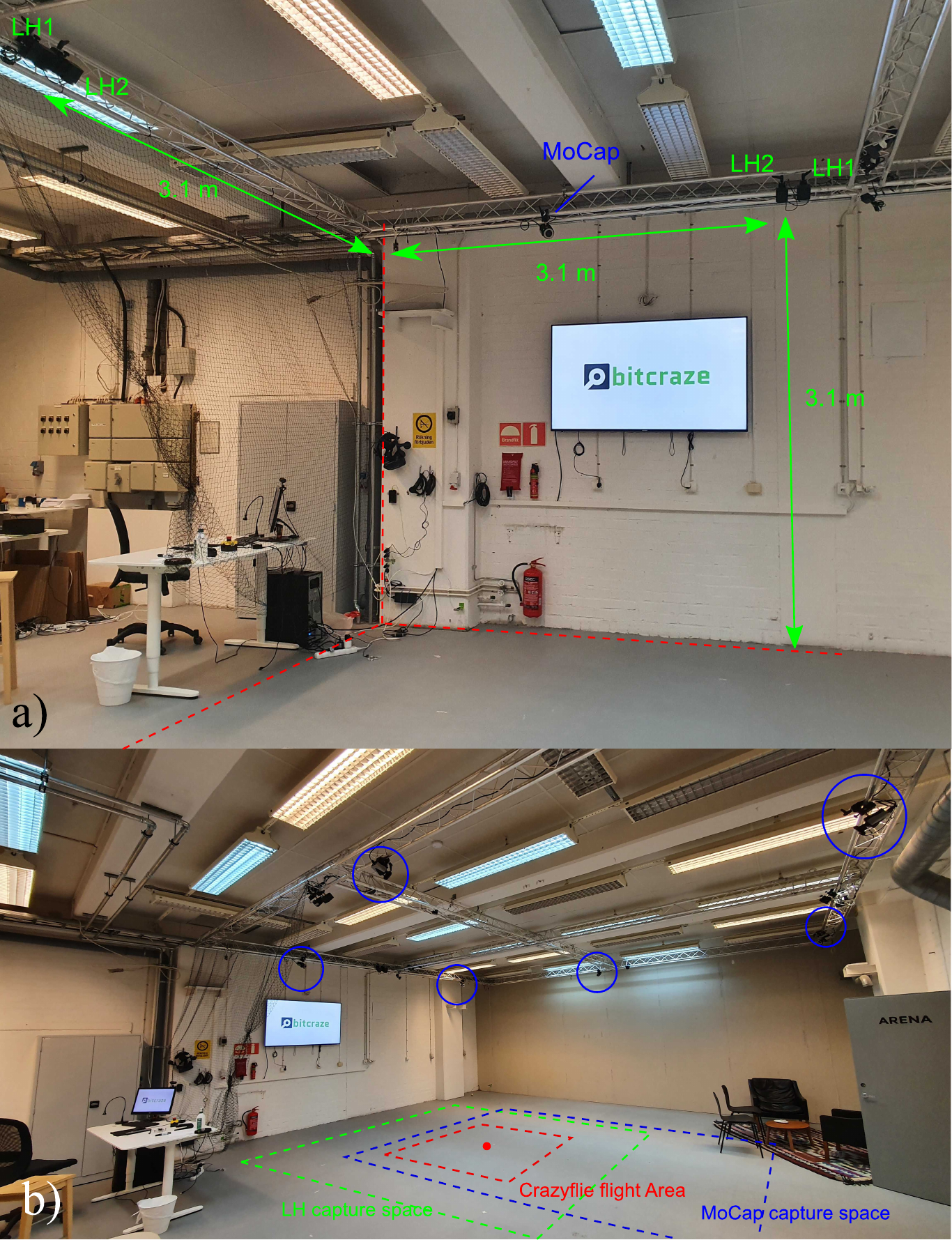}
    \caption{Photos and indications of a) the closeup of the Lighthouse setup in the flight area and b) of the full flight area with Qualisys motion capture (MoCap) cameras.}
    \label{fig:flightarea}
\end{figure}

\subsection{Flight Area and Hardware Setup}
All experiments are conducted in a \SI{7 x 7 x 3}{m} flight space that is equipped with a motion capture system (6 Qualisys Miqus M3)\footnote{\url{https://www.qualisys.com/cameras/miqus/}} and has no natural light that might interfere with the LH or motion capture (\cref{fig:flightarea}). Note that the Crazyflie flight area (\SI{1.5 x 1.5 x 1.5}{m})  was placed such that it flew in the middle of the LH's systems field of view and that it was measurable by the motion capture system for ground truth.

Both LH and motion capture operate using infrared light, which makes using both of them simultaneously difficult and has motivated previous studies to use an industrial serial manipulator to generate ground truth data rather than a motion capture system~\cite{DBLP:journals/access/IkbalRZ21}.
When using a motion capture system in its default mode, where an IR ring around the cameras is active, the LH sensors cannot perceive the beams from the LH base stations, as they use the same spectrum.
We found that this interference can be avoided entirely by using active markers: instead of using the classical retro-reflective markers, we rely on IR LEDs as markers and disable the IR ring around the motion capture cameras.
As long as the radiation cone of the IR LEDs does not include the LH sensors, the LH perception is unimpaired.
\begin{figure}[hbt!]
    \centering
    \includegraphics[width=\linewidth]{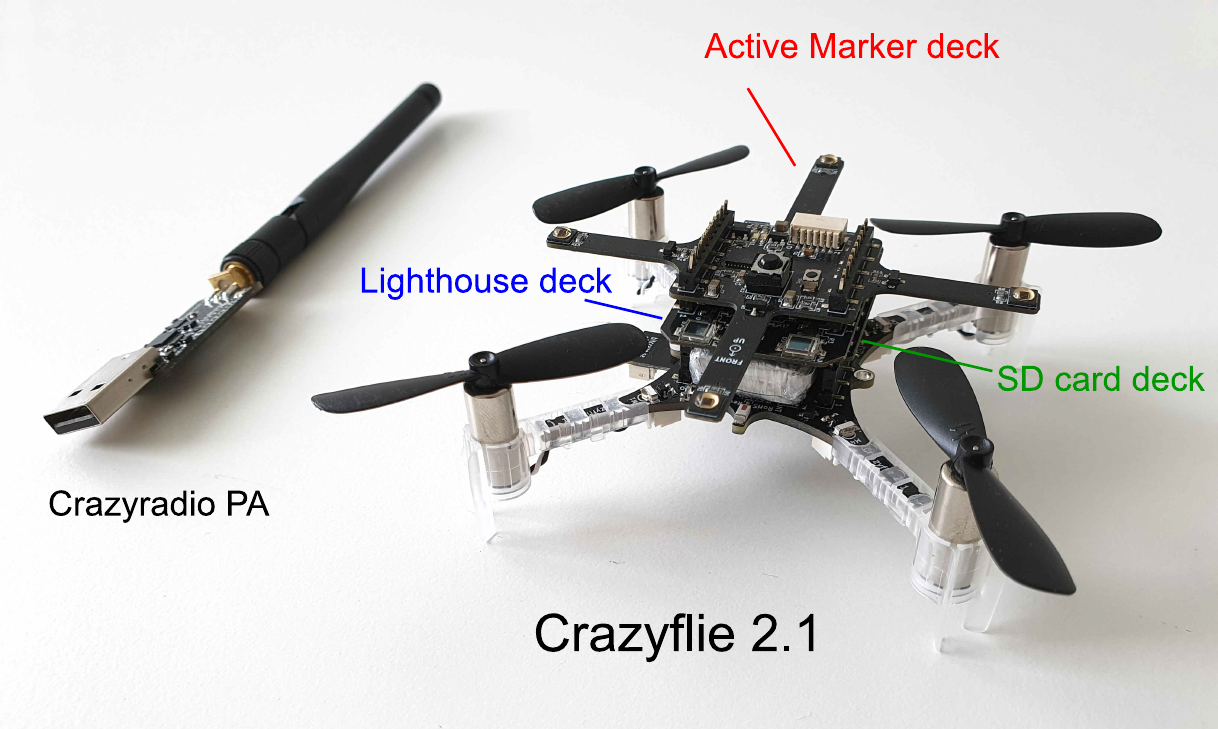}
    \caption{Picture of the Crazyflie used for data collection with (top to bottom) an active marker deck, a Lighthouse deck and a uSD-card deck, which were kept together with long male pin headers.}
    \label{fig:crazyfliesetup}
\end{figure}

In practice, we use the following commercially available off-the-shelf hardware\footnote{\url{https://www.bitcraze.io}}.
A Crazyflie 2.1 with the extension boards (so-called ``decks'') mounted on top: SD-card deck, Lighthouse deck, and an active marker deck (Fig.~\ref{fig:crazyfliesetup}).
The active marker deck has to be above the LH deck to keep the IR LED radiation cone outside the LH sensors.
The micro SD card deck is used to log the raw data that we receive onboard the Crazyflie in a space-efficient, binary encoding, where each datapoint is annotated with a high-precision microsecond timestamp.
The Crazyflie communicates with a PC over a custom radio (Crazyradio PA) only for high-level commands, such as start/stop of logging, as the radio bandwidth is insufficient to transfer the data in real-time.

\subsection{Data Collection Preparation}
Before conducting experiments, we calibrate the motion capture system using the vendors calibration kit and software.
We estimate the pose of the two base stations using the CfClient software, which computes the poses after placing the Crazyflie at the origin using \texttt{solvePnP} in OpenCV.\footnote{\url{https://docs.opencv.org/4.5.1/d9/d0c/group__calib3d.html}}
Since this method only relies on four points in one plane, it is expected to be less accurate than a typical motion capture calibration that relies on hundreds of points.
We do not attempt to align the coordinate systems, as this would not be possible with high precision.
Instead, we propose to compute the transformation in postprocessing, as described in \cref{sec:analysis}.

\begin{table}[]
    \centering
    \caption{Dataset scenarios. C.B. stands for the Crossing Beam method.}
    \begin{tabular}{c|c|c||c|c}
    LH & Estimation & Setting & \# Datasets & $\sum$ Duration [s]\\
    \hline
    \hline
    LH1 & C.B. & Stationary     & 5 & 59\\
    LH1 & C.B. & Ext. Motion    & 5 & 603\\
    LH1 & C.B. & Flight         & 5 & 416\\
    LH1 & EKF  & Stationary     & 5 & 59\\
    LH1 & EKF  & Flight         & 4 & 436\\
    LH2 & C.B. & Stationary     & 5 & 59\\
    LH2 & C.B. & Ext. Motion    & 5 & 603\\
    LH2 & C.B. & Flight         & 4 & 440\\
    LH2 & EKF  & Stationary     & 5 & 59\\
    LH2 & EKF  & Flight         & 4 & 437\\
    \end{tabular}
    \label{tab:scenarios}
\end{table}

We consider all valid combinations of LH hardware version (LH1 and LH2), state estimation method (crossing beam and EKF), and scenarios (stationary, external motion, flight), see \cref{tab:scenarios}.
For a description of the hardware and state estimation, see \cref{sec:LH}.
In the stationary scenario we place the Crazyflie at different positions and collect data for 10 consecutive seconds.
This is in particular useful to compute an estimate of the noise of the estimation, e.g., jitter.
In the external motion or movement scenario, we mount the Crazyflie on a long rod and move it around manually inside the capture space, similar to how a calibration wand is moved.
Here, higher velocities might be reached and there is no known dynamics model that can be applied.
We use various different sweeping patterns as well as movement velocities.
For the flight scenario, we fly the Crazyflie using the LH system as feedback using a sweeping pattern and randomly sampled position setpoints with two desired velocities (\SI{0.25}{m/s} and \SI{0.5}{m/s}).
Since the EKF requires a known dynamics model, the combination of EKF and external motion is not considered.

We operate the motion capture system at a fixed sample frequency of \SI{300}{Hz}.
Since we only consider positions and not orientation for this work, we do not use the rigid body tracking integrated in the Qualisys software, but instead compute the mean of the position, $\pmb p^\MC$, of the four 3D markers created by the active marker deck (the geometry of the deck is such that the mean reflects the geometric center). If there were not exactly 4 markers detected, we report NaN. Each datapoint is timestamped with the Qualisys camera timestamp, $t^\MC$, and the resulting time series 
\begin{equation}
    \mathcal D^\MC = \{(t_i^\MC, \pmb {p}_i^\MC) \,|\, i = 1,\ldots \}
\end{equation}
is written to a \texttt{mocap<id>.npy} file.

On the Crazyflie, we record the gyroscope and acceleration at \SI{100}{Hz}, the raw LH angles and the estimated position (by either the EKF or crossing beam method) as event streams. Each datapoint is timestamped with the onboard microsecond timer and the resulting time series is written to a \texttt{log<id>} file.
For brevity, we only introduce notation for the estimated position events:
\begin{equation}
    \mathcal D^\CF = \{(t_i^\CF, \pmb p_i^\CF) \,|\, i = 1,\ldots \}.
\end{equation}
All files are compact binary representations and we provide Python scripts to parse and analyze the data.

In order to enable synchronization between the motion camera clocks and the Crazyflie clock, we record the onboard microsecond timestamp when the IR LEDs on the active marker deck are turned on and off, respectively. These two events are sufficient to synchronize the clocks (offset and drift).

\section{Data Analysis}
\label{sec:analysis}

We now discuss how the collected data can be post-processed for analysis and the performance metrics that we are considering in this paper.

\subsection{Spatiotemporal Alignment}

The dataset contains data from two different clock sources (motion capture clock and Crazyflie clock) and two different coordinate systems (motion capture and LH).
For any quantitative analysis we need to align the data in both space and time.

For the temporal alignment, let $t_s^\CF$ and $t_f^\CF$ be the time of the Crazyflie clock when the IR LEDs on the active marker deck were enabled and disabled, respectively.
Let $t_s^\MC$ and $t_f^\MC$ be the time of the motion capture clock when 3D markers were first and last detected, respectively.
We can then rescale the time for each recorded Crazyflie event $i$ to
\begin{equation}
\hat{t}_i^\CF = (t_i^\CF - t_s^\CF) \frac{t_f^\MC - t_s^\MC}{t_f^\CF - t_s^\CF}.
\end{equation}
The compensation for the clock drift is important, as the Crazyflie does not have a high-precision crystal.

Next, we compute the matching ground truth position, in the motion capture reference frame, for each Crazyflie event $i$ using linear interpolation:
\begin{equation}
\hat{\pmb{p}}_i^\MC = \operatorname{interp}(\hat{t}_i^\CF, \{t_i^\MC - t_s^\MC \,|\, \forall i\}, \{\pmb{p}_i^\MC \,|\, \forall i\}).
\end{equation}
Since the sampling rate of the motion capture is roughly one order of magnitude higher than the LH events, this is a good approximation.
Note that $\hat{\pmb{p}}_i^\MC$ may be NaN, if there was no motion capture data available around time $\hat{t}_i^\CF$.

We now have a temporally aligned dataset $\tilde{\mathcal{D}}=\{(\hat{t}_i^\CF, \pmb{p}_i^\CF, \hat{\pmb{p}}_i^\MC) \,|\, i = 1,\ldots\}$, where the two positions come from a different reference frame and the transformation between the reference frames is unknown.
It is easy to estimate the transformation numerically given this aligned dataset.
We rely on an approach that uses singular value decomposition (SVD)~\cite{DBLP:journals/pami/ArunHB87} and that has been shown empirically to produce high-quality and robust results~\cite{DBLP:journals/mva/EggertLF97}.
The approach requires as only input $\tilde{\mathcal{D}}$ (ignoring the temporal component) and outputs a rotation matrix $R$ and translation vector $\pmb t$, which transforms points from the LH frame to the motion capture frame.
Finally, we can compute our spatiotemporally aligned dataset
\begin{equation}
    \begin{split}
    \hat{\pmb{p}}_i^\CF = R \pmb{p}_i^\CF + \pmb{t} \quad \forall i\\
    \hat{\mathcal{D}}=\{(\hat{t}_i^\CF, \hat{\pmb{p}}_i^\CF, \hat{\pmb{p}}_i^\MC) \,|\, i = 1,\ldots\}.
    \end{split}
\end{equation}

In practice, we found that the motion capture system has some additional latency of unknown source.
To mitigate this, we consider different offsets $t_s^o$ and $t_f^o$ in the range of $\pm \SI{100}{ms}$ and repeat the spatiotemporal alignment procedure described earlier.
We then pick the offsets that minimize the Euclidean error $\sum_i \|\hat{\pmb{p}}_i^\CF - \hat{\pmb{p}}_i^\MC\|_2$.

\subsection{Data Filtering}

The dataset contains periods of time where either no LH data was received or the Crazyflie could not be observed with the motion capture system.
The first one largely occurs during the external motion, since the LH capture space is smaller than the motion capture space.
Other reasons for missing data include interference between the two base stations of LH2 that occurs periodically, as well as a partial occlusion of the LH sensors due to the active marker deck mounted on top.
Missing data from the motion capture system occurs if not all 4 markers are visible.
Thus, we exclude datapoints where we had no motion capture ground truth. 

For the crossing beam estimation, we exclude datapoints where we did not receive both light planes from both base stations on all four sensors (by design of the crossing beam method).
We also exclude datapoints where the crossing beam $\delta$ is greater than \SI{0.1}{m}.
This variable is computed onboard as the error of the ray intersection approximation.

For the EKF estimation, we exclude datapoints where we did not receive at least one light plane for at least one LH sensor from both base stations. We noticed that the filter tends to drift during these time periods, which skews the results otherwise and makes them difficult to compare to the crossing beam method that relies on much more aggressive data filtering.

\subsection{Performance Metrics}

We are interested in quantifying precision and accuracy of the LH system. For precision, we compute the root mean square of the position (also often termed as jitter):
\begin{equation}
    P = \sqrt{\frac{1}{|\hat{\mathcal{D}}|} \sum_{i=1}^{|\hat{\mathcal{D}}|-1} \|\hat{\pmb{p}}_i^\CF - \hat{\pmb{p}}_{i+1}^\CF\|_2^2}.
\end{equation}

For the accuracy, we consider the Euclidean error for each datapoint 
\begin{equation}
    A_i = \|\hat{\pmb{p}}_i^\CF - \hat{\pmb{p}}_i^\MC\|_2,
\end{equation}
where interesting statistics include the mean $\bar A = \frac{1}{|\hat{\mathcal{D}}|} \sum_i A_i$ and maximum error $A_{\text{max}} = \max_i A_i$.

\section{Results}
\label{sec:results}

We first analyze the precision and sample frequency by considering the data in the stationary setting, see \cref{tab:precision}.


\begin{table}[]
    \centering
    \caption{Precision/Jitter and Sample Frequency.}
    \begin{tabular}{c||c|c||c|c||c}
                    &   \multicolumn{2}{c||}{LightHouse 1} & \multicolumn{2}{c||}{LightHouse 2} & MoCap\\
                    &   C.B. & EKF & C.B. & EKF & \\
        \hline
        \hline
        Freq. [Hz]  &   $30 \pm 2.4$ & N.A. & $34 \pm 18$ & N.A.& $300 \pm 0.1$\\
        Jitter [mm] &   $0.6$ & $3.9$ & $0.3$ & $0.7$ & $0.1$
    \end{tabular}
    \label{tab:precision}
\end{table}

\begin{figure}
    \centering
    \includegraphics[width=\linewidth]{images/error_boxplot.pdf}
    \caption{Accuracy of Lighthouse 1 and 2 in different scenarios. LH2 achieves better accuracy compared to LH1 in all cases. The flight scenarios have a better accuracy due to the lower velocity and focus on the volume with excellent LH coverage. EKF and crossing beam state estimators achieve a similar accuracy during our flight tests.}
    \label{fig:accuracy}
\end{figure}

For the crossing beam state estimation, the sample frequency is around \SI{30}{Hz}, which is sufficient for non-aggressive flight maneuvers.
However, in the LH2 case, this frequency has a very large standard deviation which is caused by the periodic interference of the two base stations.
An initial investigation showed that this is mostly caused by overly conservative filtering of the raw sensor data and can be improved with a software upgrade in the future.
Note that the provided dataset contains the raw sensor data as well and could be used to test less conservative filtering techniques.
For the EKF estimation, the sample frequency is arbitrary and data from LH is considered in the filter as soon as it is observed.

The precision is significantly lower for LH2 compared to LH1.
Moreover, the precision is worse when using the EKF compared to the crossing beam state estimation for both LH1 and LH2.
This indicates that the jitter induced by the IMU measurements or by the process noise is higher than the jitter induced by the LH. 
Tuning the EKF, for example with the use of this dataset, might improve the result.

The accuracy in various settings is shown using box and whisker plots in \cref{fig:accuracy}.
The mean and median Euclidean error is in all cases in the low centimeter (\SIrange{2}{4}{cm}) range.
There are some outliers, but during flight they are in the worst case about \SI{5}{cm}.
Larger outliers can be observed for the external motion case (up to \SI{18}{cm}) with the crossing beam state estimator.
These are caused by faster motions as well as operation close to the boundary of the volume that is observable by the LH base stations.
Overall, LH2 has a higher accuracy compared to LH1 in all scenarios, which we attribute to the simplified mechanics or improved factory calibration.
The two state estimators, EKF and crossing beam, have a similar accuracy during our flight tests.

\section{Conclusion}

We investigate the Lighthouse positioning system for UAV research in comparison to a traditional motion capture system by providing a dataset and analysis of accuracy and precision.
We quantify the precision in terms of RMS jitter, which is in most of our settings in the sub-millimeter range (about 5 times larger than the reference motion capture system).
We quantify the accuracy in terms of Euclidean error, which is about \SI{1}{cm} in the average case with outliers up to \SI{5}{cm} during flight, and significantly higher outliers when moving freely.
Our results indicate that the LH system is a versatile alternative to traditional motion capture systems: as feedback during flight (EKF mode) or as a ground truth (crossing beam mode) for other, less accurate positioning systems such as ultra-wideband (UWB) based systems.
The Lighthouse system's price and fully distributed operation make it in particular interesting for classroom use and distributed swarm research.

The dataset also opens up many exciting avenues for future research.
First, it would be interesting to improve the automatic metric of the crossing beam method to filter out bad state estimates.
We found that the current approach of using the $\delta$ variable does not always correlate with high errors.
Second, the precision measurements showed that LH2 has a high standard deviation for the sample frequency, which is likely caused by pessimistic filtering of the raw data.
Third, the high precision might indicate that a higher accuracy is possible as well, if better calibration methods are developed.
We believe that our dataset can be used for all of these directions as well as for tuning the EKF, as it contains the relevant raw data.



\printbibliography

\end{document}